\documentclass{article}
\usepackage{spconf,amsmath,graphicx}
\usepackage{amssymb}
\usepackage{cite}
\usepackage{booktabs}
\usepackage{multirow}
\usepackage{amsfonts} 
\usepackage[table]{xcolor}
\usepackage{times}
\usepackage{caption}
\usepackage{helvet}
\usepackage{courier}
\usepackage[hyphens]{url}
\usepackage{bibentry}
\usepackage{hyperref}

\title{QuadBox: Accelerating 3D Gaussian Splatting with Geometry-Aware Boxes}

\name{\begin{tabular}[t]{@{}c@{}}
Xinze Li$^{1}$, Bohan Yang$^{1}$, Pengxu Chen$^{3}$, Yiyuan Wang$^{1,2}$ \\
Hongcheng Luo$^{4}$, Wentao Cheng$^{1}$, Weifeng Su$^{1,5}$
\end{tabular}}

\address{\begin{tabular}[t]{@{}c@{}}
$^{1}$ Beijing Normal-Hong Kong Baptist University \quad
$^{2}$ Hong Kong Baptist University \\
$^{3}$ Jilin University \quad
$^{4}$ Xiaomi \\
$^{5}$ Guangdong Provincial Key Laboratory of Interdisciplinary Research and Application for Data Science \\
Corresponding author: Wentao Cheng
\end{tabular}}

\begin{document}
%
\maketitle
%

\begin{abstract}
\label{sec:abs}
3D Gaussian Splatting (3DGS) has emerged as an advanced technique for real-time novel view synthesis by representing scene geometry and appearance using differentiable Gaussian primitives. However, efficiently computing precise Gaussian–tile intersections remains a critical task in the rasterization pipeline. To this end, we propose QuadBox, a method that leverages four axis-aligned bounding boxes to tightly encapsulate projected Gaussians in a discrete manner. First, we derive a geometry-aware stretching factor that enables the construction of a tile-aligned QuadBox, which covers the elliptical projection and largely excludes irrelevant tiles. Second, we introduce QPass, a single-pass tile traversal algorithm that exhaustively exploits the discrete nature of QuadBox, ensuring that the tile intersection check is performed with simple interval tests. Experiments on public datasets show that our method accelerates the rendering speed of 3DGS by 1.85$\times$. Code is available at \href{https://github.com/Powertony102/QuadBox}{https://github.com/Powertony102/QuadBox}.
\end{abstract}


\section{Introduction}
\label{sec:intro}
Novel View Synthesis (NVS) aims to generate photorealistic views from novel perspectives, given only a sparse set of captured images. With its ability to combine accurate scene reconstruction and real-time performance, 3D Gaussian Splatting (3DGS) has rapidly become a cornerstone technique in novel view synthesis \cite{mildenhall2021nerf,kerbl20233d}. On one hand, leveraging the closure properties of elliptical Gaussians under affine transforms and convolution \cite{zwicker2002ewa}, 3DGS accelerates rendering by stacking 2D Gaussians for alpha compositing. On the other hand, the optimized rasterizer leverages the parallelism of modern GPUs to accelerate rendering. Nonetheless, recent efforts aim to further improve 3DGS performance in the face of diverse scene complexities and hardware constraints \cite{yu2024mip,ye2024absgs,fridovich2022plenoxels,lu2024scaffold}.

One straightforward method to accelerate rendering is pruning, which reduces the number of Gaussians in the scene \cite{fan2024lightgaussian, fang2024mini, lee2024compact, zhang2024lp, mallick2024taming, zhang2024pixel, rota2024revising}. Although pruning reduces computational complexity, it usually undermines visual fidelity, with fewer Gaussians potentially leading to information loss. Another promising paradigm involves identifying redundancies within the rasterizer pipeline to reduce unnecessary computations \cite{feng2025flashgs, wang2024adr}. In the 3DGS rasterizer, a tile-based design is adopted to enable high parallelism for processing image patches. The initial \texttt{preprocess} kernel calculates which tiles intersect with a 2D elliptical Gaussian, and the final \texttt{render} kernel uses alpha compositing to assign color to each pixel within intersecting tiles. A key limitation of the original rasterizer lies in its reliance on relaxed bounding boxes, which often include numerous tiles that receive no actual splats during the \texttt{render} kernel phase.

\begin{figure}[t] 
\centering 
\includegraphics[width=0.36\textwidth]{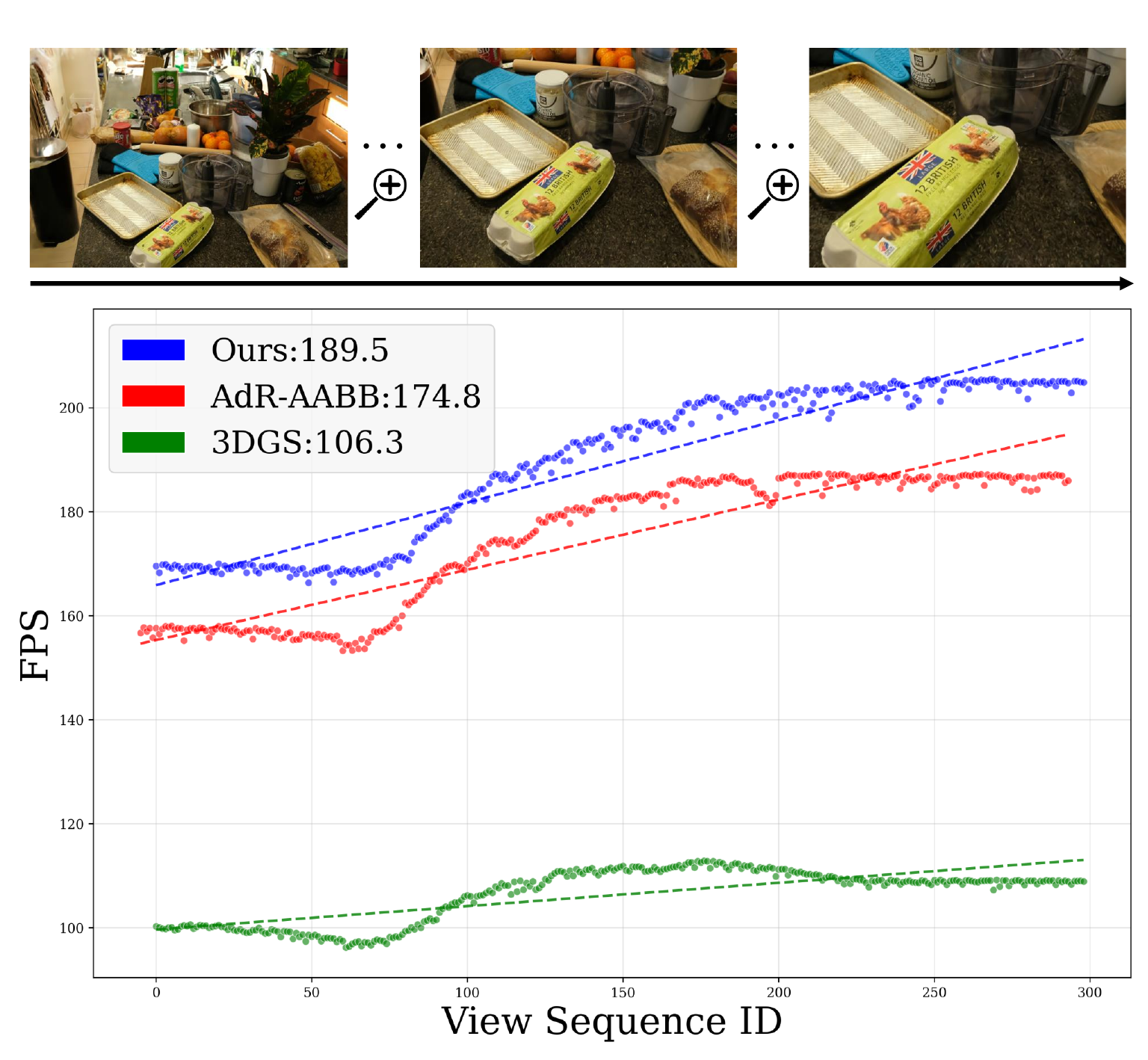} 
\caption{Rendering speed comparison during zoom-in on an indoor scene. Our method consistently outperforms the Gaussian-tile intersection algorithm (AABB) in AdR-Gaussian \cite{wang2024adr} and 3DGS in FPS.}
\label{Fig.teather} 
\end{figure}

The tile redundancy can be alleviated by employing tighter bounding boxes that account for opacity, as demonstrated in AdR-Gaussian \cite{wang2024adr}. However, its single bounding box design fails to fully capture the anisotropic shape and orientation of the ellipse, which often leads to residual redundancy along the minor axis where tiles may still fall outside the true support region. To address this problem efficiently, we introduce QuadBox, a Gaussian–tile intersection method that strives to fit the elliptical Gaussian using fundamental shapes. The method begins by approximating the elliptical footprint with four axis-aligned bounding boxes, symmetrically positioned around the Gaussian center. These sub-boxes are adaptively stretched to reflect the ellipse's eccentricity and orientation, ensuring tight but conservative coverage. The QuadBox construction is executed once per Gaussian and remains decoupled from tile coverage. Tile-level tests rely solely on integer comparisons, enabling minimal per-tile overhead and strong scalability with scene complexity.

A naive approach that iterates over each sub-box independently introduces redundant tile checks and warp divergence. To address this, we propose QPass, a branch-free single-pass tile traversal algorithm tailored for QuadBox. QPass computes the global tile range with simple integer divisions, and for each scanline, aggregates the active tile interval using fast min/max tests over intersecting sub-boxes. This avoids expensive per-pixel or analytic checks, yielding a deterministic, cache-friendly scan that visits each tile exactly once. Extensive evaluation on the Mip-NeRF 360 \cite{barron2022mip}, Tanks \& Temples \cite{knapitsch2017tanks}, and Deep Blending \cite{hedman2018deep} datasets demonstrates that our method achieves an average over 1.8$\times$ speedup compared to 3DGS \cite{kerbl20233d}.

\section{Preliminary}
\label{sec:bg}

3D Gaussian Splatting (3DGS) \cite{kerbl20233d} represents scenes via a set of Gaussian primitives ${\{ \mathcal{G}_i \}}_{i=1}^N$. Each primitive is parameterized by a center $\mu_i$, a 3D covariance $\Sigma_i$, opacity $o_i$, and spherical harmonic coefficients for color $c_i$. The geometry is defined as:
\begin{equation}
    \mathcal{G}_i(x) = \exp\left({-\frac{1}{2} (x - \mu_i)^\top {\Sigma}_i^{-1}(x - \mu_i)}\right).
\end{equation}
To render novel views, 3D Gaussians are projected into 2D space via affine transformation \cite{zwicker2002ewa}. The splatted opacity at pixel $u$ is given by $\alpha_{i} = o_{i} \mathcal{G}_i^{2D}(u)$. Pixel color $C(u)$ is then computed by accumulating contributions in front-to-back order:
\begin{equation}
    C(u) = \sum_{i=1}^{N} T_i(u)\,\alpha_i\,c_i, \quad  T_i(u) = \prod_{j=1}^{i-1}(1 - \alpha_j).
\end{equation}

The tile-based rasterizer partitions the image plane into $16 \times 16$ tiles and executes four key stages. 
1) \texttt{Preprocess} operates in a Gaussian-parallel manner, projecting primitives and assigning screen-space bounding boxes with side length $6 \sqrt{\lambda_{\text{max}}}$, where $\lambda_{\text{max}}$ is the maximum eigenvalue of the 2D covariance. 
2) \texttt{DuplicateWithKeys} maps Gaussians to their overlapping tiles. 
3) \texttt{SortPairs} orders these tile-Gaussian pairs by depth. 
4) \texttt{Render} performs per-pixel alpha blending.

Notably, the bounding boxes used in \texttt{preprocess} are conservative, often covering tiles where the Gaussian contribution is negligible. These false positives propagate through the pipeline, incurring unnecessary memory and sorting overheads. To address this, we propose QuadBox, a method that rejects non-contributing tiles during the \texttt{preprocess} stage to enhance rendering throughput.
\section{Method}
\label{sec:rw}
We introduce QuadBox, a geometric culling method designed to address tile over-approximation. By enclosing each Gaussian with four adaptive axis-aligned bounding boxes (AABBs), QuadBox ensures tight coverage of arbitrary ellipses. This structure serves as the geometric foundation for QPass, our efficient interval-based tile traversal algorithm.

\subsection{QuadBox Construction}
\label{sec:construct}
\begin{figure*}[htbp] 
\centering 
\includegraphics[width=0.80\textwidth]{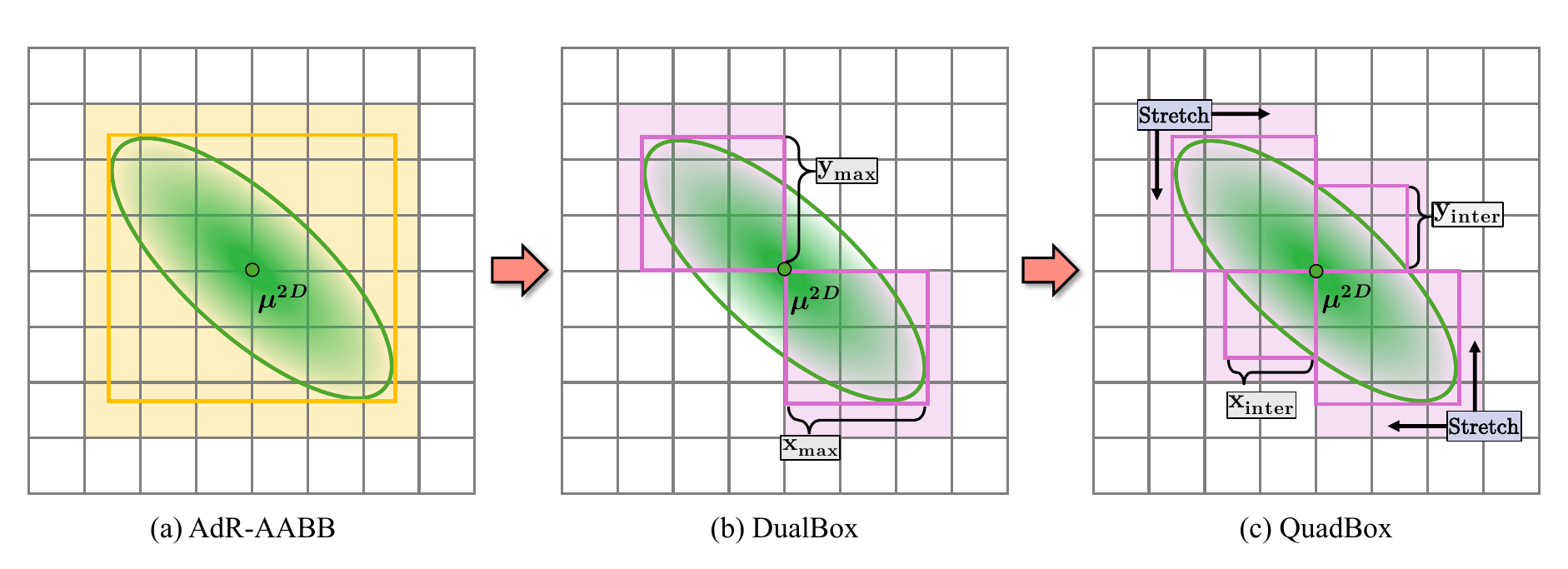} 
\caption{Evolution of QuadBox. (a) AdR-AABB \cite{wang2024adr}. (b) DualBox splits the region along the major axis. (c) QuadBox adds adaptively stretched sub-boxes to cover the minor axis, derived via closed-form geometric analysis.}
\label{Fig.quadbox} 
\end{figure*}

The construction proceeds in three phases: initialization via opacity filtering, major-axis alignment (DualBox), and adaptive stretching (QuadBox), as shown in Figure \ref{Fig.quadbox}.

\subsubsection{Initialization} 
 To ensure numerical stability during rendering, 3DGS typically ignores Gaussians whose splatted opacity falls below a predefined threshold $\alpha_\text{min}$. Leveraging this observation, we follow \cite{wang2024adr} to filter Gaussians with opacity $\alpha < \alpha_\text{min}$ ($1/255$). This constraint establishes the confidence inequality:
\begin{equation}
    2 \ln (\frac{o_i}{\alpha_\text{min}}) \leq (u - \mu_i^{2D})(\Sigma_i^{2D})^{-1} (u - \mu_i^{2D})^\top. \label{eq:alpha_min}
\end{equation}
Letting $\Lambda = (\Sigma^{2D})^{-1} = [a, b; b, c]$ and centering coordinates as $(x_d, y_d) = u - \mu_i^{2D}$, we reformulate this as the ellipse equation used in subsequent derivation:
\begin{equation}
    F(x_d, y_d) = ax_d^2 + 2bx_dy_d + cy_d^2 - \gamma \leq 0,
    \label{eq:ellipse-def}
\end{equation}
where $\gamma$ represents the threshold derived from Eq. \ref{eq:alpha_min}. This region is initially bounded by the tight AdR-AABB \cite{wang2024adr}.


\subsubsection{DualBox Extraction}
Standard AABBs introduce significant redundancy for elongated ellipses, particularly under diagonal orientations (Fig. \ref{Fig.quadbox}(a)). To mitigate this, we extract a \textit{DualBox} by retaining only the two quadrants of the AdR-AABB aligned with the ellipse's major axis. This strategy isolates the primary geometric structure, effectively suppressing empty tile coverage along the minor axis.

In the third phase, we aim to tightly enclose the parts of the ellipse not yet covered by the DualBox. To simplify the analysis, we define a local coordinate system by shifting the origin to the center of a ellipse $\mu^{2D}$. As shown in Figure \ref{Fig.quadbox}(c).


Our essential task can be rephrased as determining the intersection points between the ellipse and y-axis in the centered coordinate. The stretching factor we need is the ratio between the distance from the intersection points to $\mu_y$, which can be denoted as $y_{\text{inter}}$ and $y_{max}$ of the ellipse. To compute $y_{\text{inter}}$, we simply set $x_d = 0$ in the ellipse definition. As a result, we can construct the stretching factor as:

\begin{equation}
    f = \frac{y_{\text{inter}}}{y_{\text{max}}} = \frac{x_{\text{inter}}}{x_{\max}}
\end{equation}


Setting $x_d = 0$ and \(y_d=0\) in Equation \ref{eq:ellipse-def} easily gives:
\begin{equation}
    c\,y_d^2=\gamma\Rightarrow y_{\text{inter}}=\sqrt{\gamma/c},
    \qquad
    a\,x_d^2=\gamma\Rightarrow x_{\text{inter}}=\sqrt{\gamma/a}
\end{equation}

Critically, we emphasize the importance of crafting not only a straightforward but also high numerical precision formulation that steers clear of excessive use of $\sqrt{\cdot}$ and numerous $\times$ or $\div$ operations. These operands inherently inflate computational overhead and hinder \texttt{preprocess} efficiency. In the Appendix, we present a comparative analysis of the computational cost associated with various solution strategies; based on this study, we selected the current closed-form expression that offers the optimal balance between precision and efficiency.

Formally, we can rewrite \ref{eq:ellipse-def} first by:

\begin{equation}
    u^\top(\Sigma^{2D})^{-1}u = \gamma .
\end{equation}

For simplicity, here we use $\Lambda \in \mathbb{R}^{2 \times 2}$ to denote $(\Sigma^{2D})^{-1}$. Also, $\Lambda$ is notably a positive definite matrix. 

For any direction \(v\in\mathbb{R}^2\), the maximal support of the ellipse $ \max\{v^\top u:\;u^\top C u=\gamma\} $ is

\begin{equation}
\begin{aligned}
    \max v^\top u &= \sqrt{\,\gamma\, v^\top \Lambda^{-1} v\,}, \\
    u^\star &= \frac{\sqrt{\gamma}}{\sqrt{v^\top \Lambda^{-1} v}}\, \Lambda^{-1}v,
\end{aligned}
\label{eq:support}
\end{equation}

which follows either from the Lagrangian first-order condition or from Cauchy--Schwarz after a Cholesky factorization of \(\Lambda\). Detailed proof can be found in our appendix.

Choosing \(v=e_y=(0,1)^\top\) and \(v=e_x=(1,0)^\top\) yields

\begin{equation}
    y_{\max}=\sqrt{\,\gamma\,(\Lambda^{-1})_{22}\,},
    \qquad
    x_{\max}=\sqrt{\,\gamma\,(\Lambda^{-1})_{11}\,}.
    \label{eq:extents-inv}
\end{equation}

For a \(2\times2\) symmetric matrix,
\begin{equation}
    \Lambda^{-1}=\dfrac{1}{ac-b^2}
\begin{bmatrix}
c & -b\\ -b & a
\end{bmatrix}
\end{equation}

hence
\begin{equation}
    y_{\max} \;=\; \sqrt{\frac{\gamma\,a}{ac-b^2}},
    \qquad
    x_{\max} \;=\; \sqrt{\frac{\gamma\,c}{ac-b^2}}.
    \label{eq:extents-closed}
\end{equation}

Combining \eqref{eq:extents-closed} with the axis intercepts gives the purely algebraic closed form
\begin{equation}
    \boxed{\quad
    f \;=\; \sqrt{1 - \frac{b^2}{ac}}
    \;=\; \frac{1}{\sqrt{\,c\,(\Lambda^{-1})_{22}\,}}
    \;=\; \frac{1}{\sqrt{\,a\,(\Lambda^{-1})_{11}\,}} \quad}.
    \label{eq:f-closed}
\end{equation}

which depends only on the entries of \(\Lambda\) and is independent of the threshold \(\gamma\) and the center translation.
Because \(\Lambda\succ0\) implies \(a>0\), \(c>0\), and \(ac-b^2>0\),
the expression satisfies \(0<f\le 1\), with equality \(f=1\) for circles or when the principal axes align with the coordinate axes.

The reduced tiles from original 3DGS lead to a shortened training and rendering time, also indicating a promising FPS. Moreover, the discrete formulation of QuadBox provides the structural precondition for the following described QPass algorithm, enabling exact tile intersection to be computed via simple integer range checks rather than analytic geometry.


\subsection{QPass: QuadBox-based Tile Traversal}
\subsubsection{QuadBox Stretch}
\begin{figure}[htbp] 
\centering 
\includegraphics[width=0.32\textwidth]{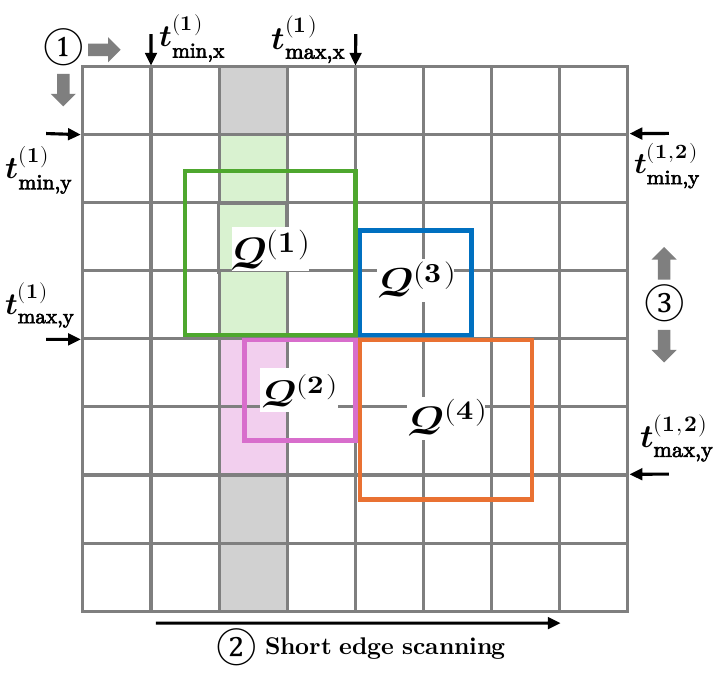} 
\caption{QPass traversal: (1) Compute global bounds. (2) Scan along the shorter axis. (3) Merge vertical intervals of intersecting sub-boxes to identify active tiles in a single pass.}
\label{Fig.quadbox_traversal} 
\end{figure}

\label{sec:qpass}
After obtaining a QuadBox for each Gaussian primitive, we need to identify the intersected tiles, thereby establishing the Gaussian-to-tile mapping. This mapping is then inverted through the \texttt{DuplicateWithKeys} step to enable subsequent tile-level parallel rendering. A naive method to compute the tile coverage of the QuadBox is by looping over the grid-aligned bounds of each sub-box with nested \texttt{for} loops, while avoiding redundant visits through conditional checks on overlapping regions. However, independent traversal of each sub-box results in repeated edge tile evaluations, while the branching complexity from variable tile intervals induces significant warp divergence. Moreover, full-grid scanning introduces numerous non-contributing tiles that are later culled by QuadBox, leading to unnecessary computation and degraded rendering performance.

We propose QPass, a single-pass tile traversal algorithm, that avoids redundant tile checks while ensuring complete coverage. At its core, QPass leverages the discrete grid structure and the tight bounds provided by our QuadBox representation. Figure \ref{Fig.quadbox_traversal} illustrates an exemplary process of how QPass works. Given a QuadBox $\mathcal{Q}$, we partition it into four quadrant-aligned sub-boxes, denoted as $\{\mathcal{Q}^{(i)} \} _{i=1}^4$. For each sub-box $\mathcal{Q}^{(i)}$, we first compute its tile-space bounds based on its 2D coordinates, denoted as  $\{t_{\min,x}^{{(i)}}, t_{\max,x}^{{(i)}}, t_{\min,y}^{{(i)}}, t_{\max,y}^{{(i)}} \}_{i=1}^4$. These bounds can be directly aggregated to obtain the global tile range of the full QuadBox. To reduce traversal overhead, we select the shorter axis of the global tile range for scanline iteration. Assuming a column-wise traversal, any tile column falling outside the global x-range is immediately skipped. Within each valid column, we identify the sub-boxes whose tile ranges intersect the current column via a fast interval overlap check. The vertical tile range is then determined by aggregating the y-extent of these intersecting sub-boxes using simple min-max operations.

\section{Experiments}

\begin{figure}[h] 
\centering 
\includegraphics[width=0.46\textwidth]{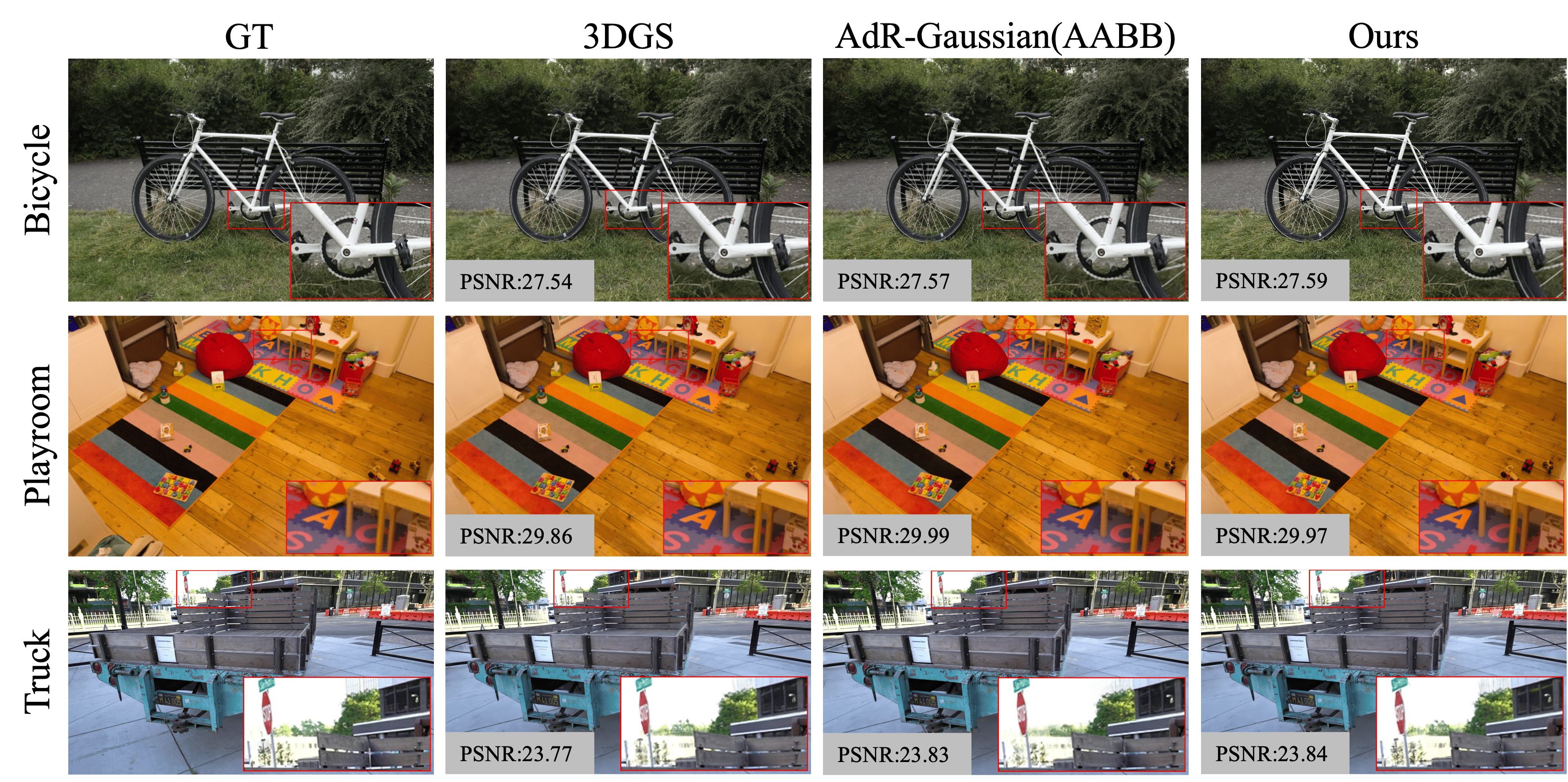} 
\caption{Qualitative comparison with 3DGS and AdR-AABB. We visualize representative scenes: \textit{bicycle} (Mip-NeRF 360), \textit{playroom} (Deep Blending), and \textit{truck} (Tanks \& Temples). Reported PSNR values indicate dataset averages.}
\label{Fig.quantative_result} 
\end{figure}

\subsection{Datasets and Implementation Details}
\label{sec:exp-dataset}

\begin{table*}[htbp]
    \centering
    \footnotesize
    \captionsetup{singlelinecheck=false}
    \caption{Comparison of acceleration strategies on NVIDIA A100. Top: Core intersection modules on vanilla 3DGS. Bottom: Plug-and-play integration with recent variants. ``3DGS + Ours'' denotes our full pipeline. The \textbf{bold} results represent the best performance of each sub-comparison.
    }
    \resizebox{2.1\columnwidth}{!}{%
    \begin{tabular}{l | c c c c c | c c c c c | c c c c c }
        \toprule
        \multirow{2}{*}{Method} 
        & \multicolumn{5}{c|}{MipNeRF-360} & \multicolumn{5}{c|}{Deep Blending} & \multicolumn{5}{c}{Tanks and Temples} \\
        \cmidrule(l{2pt}r{2pt}){2-6} \cmidrule(l{2pt}r{2pt}){7-11} \cmidrule(l{2pt}r{2pt}){12-16}
        & $\mathrm{N_{GS}}$ $\downarrow$ & PSNR $\uparrow$ & SSIM $\uparrow$ & LPIPS $\downarrow$ & FPS $\uparrow$
         & $\mathrm{N_{GS}}$ $\downarrow$ & PSNR $\uparrow$ & SSIM $\uparrow$ & LPIPS $\downarrow$ & FPS $\uparrow$
         & $\mathrm{N_{GS}}$ $\downarrow$ & PSNR $\uparrow$ & SSIM $\uparrow$ & LPIPS $\downarrow$ & FPS $\uparrow$ \\
        \midrule
        \multicolumn{16}{c}{\textbf{Gaussian-tile Intersection}} \\
        \midrule
        3DGS
            & 2.76M & 27.54 & 0.817 & 0.216 & 180
            & \textbf{2.48M} & 29.86 & 0.908 & 0.242 & 173 
            & \textbf{1.57M} & 23.77 & 0.853 & 0.170 & 214
            \\
        AdR-AABB
            & \textbf{2.73M} & 27.57 & 0.817 & \textbf{0.215} & 302
            & 2.55M & \textbf{29.99} & 0.908 & 0.242 & 348 
            & \textbf{1.57M} & 23.83 & \textbf{0.854} & 0.170 & 319
            \\
        3DGS + Ours 
            & 2.77M & \textbf{27.59} & \textbf{0.818} & \textbf{0.215} & \textbf{322}
            & 2.57M & 29.97 & \textbf{0.908} & \textbf{0.242} & \textbf{379}
            & 1.60M & \textbf{23.84} & 0.854 & \textbf{0.169} & \textbf{335} 
            \\
         \midrule
         \multicolumn{16}{c}{\textbf{Plug-and-Play Experiments}} \\
          \midrule
          DashGaussian
            & \textbf{2.26M} & 27.46 & 0.808 & 0.233 & 142
            & \textbf{1.81M} & 29.86 & 0.906 & 0.253 & 159
            & \textbf{1.17M} & 24.11 & \textbf{0.852} & 0.184 & 197 \\
        +Ours 
            & 2.37M & \textbf{27.50} & \textbf{0.811} & \textbf{0.230} & \textbf{305}
            & 1.91M & \textbf{29.83} & \textbf{0.906} & \textbf{0.251} & \textbf{331}
            & 1.20M & \textbf{24.13} & \textbf{0.852} & \textbf{0.183} & \textbf{364} \\
        \midrule
        Compact-3DGS
            & \textbf{1.43M} & 27.03 & 0.800 & \textbf{0.243} & 221
            & \textbf{1.04M} & 29.94 & 0.906 & \textbf{0.259} & 296
            & \textbf{0.84M} & 23.33 & \textbf{0.852} & \textbf{0.201} & 282 \\
        +Ours
            & 1.45M & \textbf{27.06} & \textbf{0.808} & 0.247 & \textbf{441}
            & 1.07M & \textbf{29.95} & \textbf{0.908} & 0.261 & \textbf{684}
            & 0.85M & \textbf{23.34} & \textbf{0.835} & 0.206 & \textbf{463}
        \\
        \bottomrule
    \end{tabular}
    }
    \label{tab:quantative-result}
\end{table*}

\begin{table}[htbp]
\centering
\footnotesize
\resizebox{0.95\columnwidth}{!}{%
\begin{tabular}{ l | ccc | ccc }
\toprule
    \multirow{2}{*}{Method} 
    & \multicolumn{3}{c|}{AdR-AABB} 
    & \multicolumn{3}{c}{QuadBox}  \\
    \cmidrule(lr){2-4} 
    \cmidrule(lr){5-7} 
    & $\mathrm{N_{GS}}$ $\downarrow$ & PSNR $\uparrow$ & FPS $\uparrow$
    & $\mathrm{N_{GS}}$ $\downarrow$ & PSNR $\uparrow$ & FPS $\uparrow$\\
    \midrule
    
    bicycle     & 4.89M & 25.31 & 199 
                & 5.03M & 25.23 & \textbf{206} \\
    flowers     & 2.86M & 21.61 & 360
                & 2.95M & 21.61 & \textbf{367} \\
    garden      & 4.21M & 27.51 & 241
                & 4.18M & 27.48 & \textbf{251} \\
    stump       & 4.30M & 26.64 & 309
                & 4.30M & 26.69 & \textbf{318} \\
    treehill    & 3.18M & 22.61 & 312
                & 3.31M & 22.58 & \textbf{318} \\
    \midrule
    room        & 1.35M & 31.69 & 324
                & 1.35M & 31.67 & \textbf{363} \\
    counter     & 1.10M & 29.09 & 312
                & 1.11M & 29.09 & \textbf{347} \\
    kitchen     & 1.62M & 31.61 & 242
                & 1.62M & 31.51 & \textbf{261} \\
    bonsai      & 1.10M & 32.07 & 423
                & 1.09M & 32.43 & \textbf{464}\\

\bottomrule

\end{tabular} 
}
\caption{Per-scene comparison between AdR-AABB and our proposed QuadBox method on Mip-NeRF 360 (measured with NVIDIA A100). The \textbf{bold} results show the best FPS performance.}
\label{tab:perscene}
\end{table}
We evaluate on Mip-NeRF 360 \cite{barron2022mip}, Deep Blending \cite{hedman2018deep}, and Tanks \& Temples \cite{knapitsch2017tanks}. Following standard protocols \cite{kerbl20233d}, we use COLMAP poses and sparse point clouds for initialization.
Our method is implemented as a custom differentiable rasterizer within the official 3DGS codebase. Experiments are conducted on NVIDIA A100 and RTX 4090 GPUs. Rendering throughput (FPS) is measured via CUDA kernel timing, averaged over three independent trials.

\subsection{Comparisons}
We benchmark against vanilla 3DGS \cite{kerbl20233d} and AdR-Gaussian \cite{wang2024adr} (AdR-AABB only). As shown in Table~\ref{tab:quantative-result} (top), vanilla 3DGS is limited by conservative bounding boxes (180 FPS on Mip-NeRF 360). AdR-AABB improves throughput to 305 FPS via opacity-aware pruning but remains restricted by axis alignment. Our QuadBox method leverages quadrant-aware partitioning to achieve tighter coverage, boosting rendering speed to 322 FPS while maintaining or slightly improving PSNR.
Per-scene analysis (Table~\ref{tab:perscene}) further confirms that QuadBox consistently outperforms AdR-AABB across diverse indoor and outdoor scenes, validating its robustness in handling complex geometries. Qualitative comparisons are shown in Figure \ref{Fig.quantative_result}.

Based on the results in Table~\ref{tab:perscene}, we further perform a per-scene comparison between AdR-AABB and our method to assess robustness across diverse environments. Our method consistently outperforms AdR-AABB in rendering speed across nearly all scenes. This suggests that our QuadBox-based tile pruning achieves better runtime efficiency without compromising image fidelity or requiring more Gaussians. The performance gain holds across both indoor and outdoor environments, highlighting the general robustness and scalability of our rasterization strategy. A qualitative comparison is shown in Figure \ref{Fig.quantative_result}. 

\subsection{Plug-and-Play Results}
We integrate our rasterizer into two state-of-the-art backbones to demonstrate generality (Table \ref{tab:quantative-result}, bottom). 
DashGaussian \cite{chen2025dashgaussian}: Integration yields a $>1.85\times$ speedup (peaking at 364 FPS) with consistent quality gains, proving compatibility with momentum-based densification.
Compact-3DGS \cite{lee2024compact}: Our method accelerates this memory-efficient model by $>1.64\times$ (up to 684 FPS) while improving SSIM/PSNR. These results verify that our geometry-aware culling enhances efficiency across different optimization strategies without sacrificing model compactness.

\begin{table}[htbp]
\centering
\footnotesize
\resizebox{\columnwidth}{!}{%
\begin{tabular}{l|cccccc}
\toprule
Method & $\mathrm{N_{GS}}$ $\downarrow$ & PSNR $\uparrow$ & SSIM $\uparrow$ & LPIPS $\downarrow$ & FPS $\uparrow$ \\
\midrule
3DGS & \textbf{2.74M} & 27.54 & 0.815 & 0.216 & 247  \\
3DGS+AdR-AABB & 2.78M & \underline{27.59} & \underline{0.817} & \underline{0.215} & 425 \\
3DGS+DualBox & 2.79M & 27.43 & 0.814 & 0.217 & \textbf{467}  \\
3DGS+$\text{QuadBox}^{\dagger}$ & 2.78M & \textbf{27.60} & \textbf{0.818} & \textbf{0.214} & 413 \\
3DGS+QuadBox & \underline{2.77M} & \underline{27.59} & \underline{0.817} & \textbf{0.214} & \underline{460}  \\

\bottomrule
\end{tabular} 
}
\caption{The ablation study of our proposed QuadBox (measured with NVIDIA 4090). The $\dagger$ denotes our proposed method without QPass traversal. The \textbf{bold} and \underline{underlined} results show the best and second best performance.}
\label{table:ablation_quadbox}
\end{table}

\subsection{Ablation Study}
Table~\ref{table:ablation_quadbox} evaluates module contributions on an RTX 4090. AdR-AABB \cite{wang2024adr} accelerates rendering but offers coarse coverage. \textit{DualBox} boosts FPS via aggressive culling but degrades quality due to symmetric approximation errors. The intermediate $\text{QuadBox}^{\dagger}$ recovers this quality loss, validating the necessity of auxiliary sub-boxes. Finally, our full QuadBox with QPass achieves the optimal balance, surpassing $\text{QuadBox}^{\dagger}$ in speed and outperforming AdR-AABB in both efficiency (460 vs 425 FPS) and fidelity. This confirms QuadBox as a robust, high-performance intersection strategy.
\section{Conclusion}
We present a novel tile-based rasterization strategy that rethinks how Gaussians interact with discrete grids. By introducing QuadBox, a quadrants-aware bounding scheme, and QPass, a branch-free traversal mechanism, we effectively reduce redundant tile checks while preserving full coverage. Unlike existing approaches that rely on uniform heuristics or coarse approximations, our method adapts to the geometry and anisotropy of each Gaussian, enabling precise culling with minimal overhead. Extensive experiments across high-complexity scenes demonstrate consistent acceleration over state-of-the-art baselines, with either improved or retained rendering quality. Our design is orthogonal to most existing Gaussian splatting pipelines and can be readily integrated to enhance scalability and performance under dense visibility conditions.

\section{Acknowledgement}
This work was supported by the National Natural Science Foundation of China (Grant No. 62306154).

\bibliographystyle{IEEEbib}
\bibliography{refs}

\end{document}